\documentclass{article}
\usepackage{ijcai20}

\usepackage{times}
\usepackage{soul}
\usepackage{url}
\usepackage[hidelinks]{hyperref}
\usepackage[utf8]{inputenc}
\usepackage[small]{caption}
\usepackage{graphicx}
\usepackage{amsmath}
\usepackage{amsthm}
\usepackage{booktabs}

\usepackage{epsfig}
\usepackage{comment}
\usepackage{multirow}
\usepackage{import}
\usepackage[dvipsnames]{xcolor}
\usepackage{amsfonts}
\usepackage{amssymb}
\usepackage{subfig}

\usepackage[linesnumbered,algoruled,boxed,lined]{algorithm2e}

\newtheorem{definition}{Definition}

\title{Membership Inference Attacks Against Object Detection Models
}

\author{
Yeachan Park$^1$\And
Myungjoo Kang$^1$\footnote{Contact Author}
\\
\affiliations
$^1$Department of Mathematical Science, Seoul National University, Republic of Korea\\
\emails
\{ychpark, mkang\}@snu.ac.kr
}

\begin{document}

\maketitle

\begin{abstract}

  Machine learning models can leak information regarding the dataset they have trained. In this paper, we present the first membership inference attack against black-boxed object detection models that determines whether the given data records are used in the training. To attack the object detection model, we devise a novel method named as called a canvas method, in which predicted bounding boxes are drawn on an empty image for the attack model input. Based on the experiments, we successfully reveal the membership status of privately sensitive data trained using one-stage and two-stage detection models. We then propose defense strategies and also conduct a transfer attack between the models and datasets.  Our results show that object detection models are also vulnerable to inference attacks like other models.
  
\end{abstract}

\section{Introduction}

Over the past few years, deep neural networks have been widely adopted in various computer vision tasks such as image classification, object detection, and semantic segmentation. Many deep learning models in various fields have been developed using a wide variety of data. These data often contain privately sensitive information such as medical records, personal photos, personal profiles, and financial information. 
 If designed without considering adversarial threats, the model can leak sensitive information of the dataset it has trained. In \cite{shokri}, it was demonstrated that even with black-box access, an adversary can conduct a membership inference attack that determines whether a data record is a part of the training set.
 
  Early studies on membership inference attacks have focused on classification tasks \cite{ml-leaks}. Several other adversarial attacks against the object detection model have been studied, the results of which indicate the potential leakage of the model\cite{xie2017adversarial,wei2018transferable}. Through this study, we have begun extending the membership inference attack to object detection tasks. 
  
   Datasets used in an object detection model can also be subject to privacy leaks. Examples of such data include outdoor pedestrian data, photos with sensitive text, and video data for autonomous driving. The membership inference of detection models can be helpful to assess whether data are collected illegally for training purposes, and attack vulnerability can be viewed as a gateway to further attacks.
  
 Compared to classifiers, there are difficulties in attack detection models: 1) In classification tasks, only the last logit of the same size is regarded, whereas in object detection tasks, all predictions based on the location of the objects are of concern. 2) Object detection tasks may have multiple objects in a single image, whereas a usual image classification task has a single object. To address these issues, we propose the canvas method for attacking an object detection model and tracing the differences in the views among the trained and test data. 
 
In summary, this paper makes the following contributions:

\begin{itemize}
    \item We first propose a new membership inference attack on object detection models with black-box access. We describe the proposed canvas method, which draws a predicted bounding box distribution on an empty canvas for convolutional neural network (CNN) classification networks. Using this method, we can achieve a higher performance than conventional machine learning methods on the PASCAL VOC dataset.
    \item We found experimentally that our attack method is robust to various types of object detection models. In addition, we showed that membership inference attacks are also successful on privately sensitive data with seemingly little difference between accuracy of the training and test datasets. We also conducted a transfer attack between different models and datasets.
    \item We suggest the use of defense methods applying  a differentially private algorithm. Experiment results show that the differentially private (DP) algorithm can defend against a membership inference with a calculated amount of privacy loss.

\end{itemize}

\section{Background and Related Work}

\subsection{Membership Inference Attack}
   The end of a membership inference attack is to determine whether the given data record is in the training dataset of the target model. A membership inference attack is based upon the assumption that the target model has a different view of the training data than that of test data that was not seen before. Although overfitting is considered to be a root cause of this membership disclosure, it cannot be the only cause \cite{long2018understanding}. The attack model may have black-box and white-box access to the target model. Under the white-box access scenario, the attack model has access to certain versions of input data or intermediate layers as well as trained parameters of the target model. White-box knowledge is powerful but not realistic because the target model may not provide detailed information. In a black-box setting, the attacker does not have direct access to the target model parameters. The attack model can only access the input data and the model output predictions. The attack model should identify the difference between the inferred predictions of the training and test samples of the target model. To achieve this aim, shadow models trained using the same algorithm are built on shadow datasets sampled from a similar distribution as the target datasets but do not contain the target training data. The attack model queries the shadow model and learns to distinguish whether the shadow model output comes from the training set.
   
   Shokri et al. \shortcite{shokri} first presented the first membership inference attack against machine learning models. Ahmed et al. \shortcite{ml-leaks} enhanced an attack by relaxing some of the assumptions. Hayes et al. \shortcite{hayes2019logan} describes a membership inference attack against generative models. To mitigate the risk of a membership inference, Rahman et al. \shortcite{rahman2018membership} and Nasr et al. \shortcite{nasr2018machine} designed differentially private models and devised an adversarial regularization, respectively.  

\subsection{Object Detection}

Object detection is a widely used computer vision task that deals with detecting an instance of a semantic objects in images or videos. 
There are mainly two types of methods for object detection using deep learning, namely, one-stage and two-stage detection.

\vspace{5pt} \noindent {\bf One-Stage Detection} \hspace{5pt} 
One-stage detectors such as YOLO \cite{redmon2016you} or SSD \cite{liu2016ssd}  treat an object detection problem as an end-to-end simple regression problem. The one-stage model directly predicts the class scores and bounding box coordinates concurrently.

\vspace{5pt} \noindent {\bf Two-Stage Detection} \hspace{5pt} 
 A two-stage detection model such as Faster R-CNN \cite{ren2015faster} is divided into two stages. The model first generates region proposals by narrowing down the number of possible object locations by filtering out most of the background samples on a region proposal network (RPN). The model then passes the proposals through the CNN head to classify the labels and regress the bounding boxes. 
 
\subsection{Datasets}
 \vspace{5pt} \noindent {\bf PASCAL VOC Dataset (2007,2012) \cite{everingham2010pascal}} \hspace{5pt} 
 PASCAL VOC datasets have been widely adopted as benchmark datasets in basic object detection tasks. The PASCAL VOC datasets consist of VOC2007 and VOC2012. The datasets contain 20 object categories including people, bicycles, birds, bottles, dogs, etc.

 \vspace{5pt} \noindent {\bf INRIA Pedestrian Dataset  \cite{dalal2005histograms} } \hspace{5pt} 
 The INRIA Pedestrian dataset is popular for pedestrian detection, which consists of 614 images for training and 288 images for testing.

 \vspace{5pt} \noindent {\bf SynthText \cite{gupta2016synthetic} } \hspace{5pt} 
The SynthText dataset is a synthetically generated text dataset in which several words are placed in imgaes of natural scenes. The dataset consists of approximately 800 thousand images and 8 million synthetic word instances in various languages.

\begin{figure*}[t]
\begin{center}
\includegraphics[width=0.9\linewidth,height=0.35\linewidth]{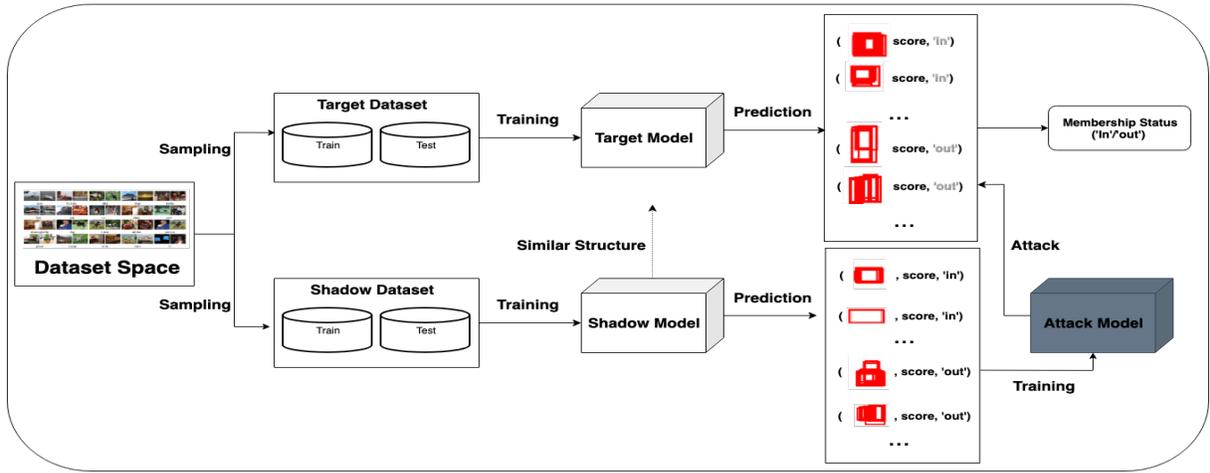}
\end{center}
   \caption{Overview of membership inference attack on object detection model. The target and shadow datasets are sampled from the same dataset space. The target model trains using its target dataset and the shadow model, which has a similar structure as the target model, trains using its shadow dataset. The predicted values of the target and shadow models are expressed as bounding boxes and their prediction scores along with their membership status labels ("in" for the training set "out" for the test set). Finally, the attack model which trains using the shadow model's prediction and membership status, attacks the target model by passing the target records, and estimates their membership status probabilities for each target example.}
\label{fig:overview}
\end{figure*}

\section{Attack Methodology}
In this section, we propose a membership inference attack for object detection models. An overview of the membership inference attack is illustrated in Figure \ref{fig:overview}.
The setting of our membership inference attack is as followes:  

 \vspace{5pt} \noindent {\bf Assumption} \hspace{5pt} 
 We assume that the adversary has black-box access to the target model. The adversary can obtain final logit values but no  other specific intermediate layer weight information of the target models. For the given target object detection model $f_{target}$ and input image sample $x_{i}$ the target model returns the proposed bounding boxes $bbox_{j}= ((x_{j}^{0},y_{j}^{0}),(x_{j}^{1},y_{j}^{1}) )$ and prediction scores $s_{j},(j=1,2,...,N_{b})$ where $ (x_{j},y_{j})$ and $N_{b}$ denote the corner of the bounding box and the number of proposed boxes, respectively. In addition, the adversary can set a score threshold $\theta_{score}$ and non-maximum suppression (NMS) thresholds(  $\theta_{nms}$ for one-stage, \{ $\theta_{nms}^{rpn}, \theta_{nms}^{head}$ \} for two-stage detectors) to customize the personal preference of the attacker. In addition, it is assumed that the target and shadow data do not overlap, i.e.,  $D^{train}_{shadow} \cap D^{train}_{target} = \varnothing $

\begin{figure}[t]
\begin{center}
\includegraphics[width=0.9\linewidth]{{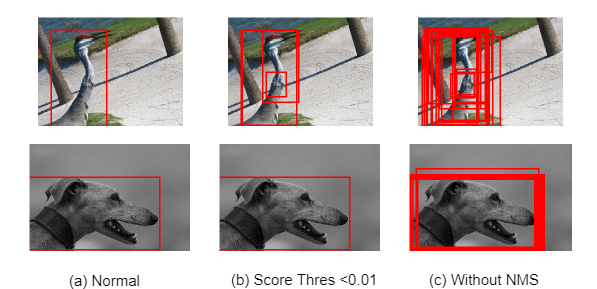}}
\end{center}
\caption{ Predicted bounding boxes in training and test examples. The first row shows the training examples and their predicted boxes. Below are test examples and their predicted boxes. 
   }
\label{fig:predicted_box_comparison}
\end{figure}

\subsection{Motivation}
  The basic idea of a membership inference attack is that the model has a different view on the trained data and unseen data. For a classification task, the model tends to achieve a  high prediction score on the training samples over the test samples. Therefore, the attack model is able to classify the membership status using the last posterior logit value of a given sample. Similarly, as shown in Figure \ref{fig:predicted_box_comparison}, the object detection model tends to achieve consistent box predictions on the training samples while showing an uncertainty regarding the test samples.

\subsection{Gradient Tree Boosting}
  Gradient tree boosting is a widely used classification algorithm for numerous applications. Specifically, we use XG-BOOST \cite{chen2016xgboost}, a popular algorithm applied feature classification, to distinguish whether a given example is in the training sample. For the predicted bounding box coordinates and prediction scores $({ bbox_{j},s_{j}} )$, we concatenate them in a long 1-D vector: $(x_{1}^{0},y_{1}^{0},x_{1}^{1},y_{1}^{1},s_{1} , ... ,  x_{N_{b}}^{0},y_{N_{b}}^{0},x_{N_{b}}^{1},y_{N_{b}}^{1},s_{N_{b}}) $, and pad them with zero values to allow all vectors to have the same length. Using these vectors, we proceed with the membership classification using XG-BOOST.
\subsection{Convolutional Neural Network Based Method}
The next method applied to the attack model is CNN based approach. An object detection task differs from a classification because the model predicts 1) the box location information and 2) the bulk of the bounding boxes, most of which may be unhelpful. Therefore, we propose a new approach, called the canvas Method, to adequately process a predicted array for a CNN-based attack model.

\begin{figure}[t]
\begin{center}
\includegraphics[width=0.9\linewidth,height=0.4\linewidth]{{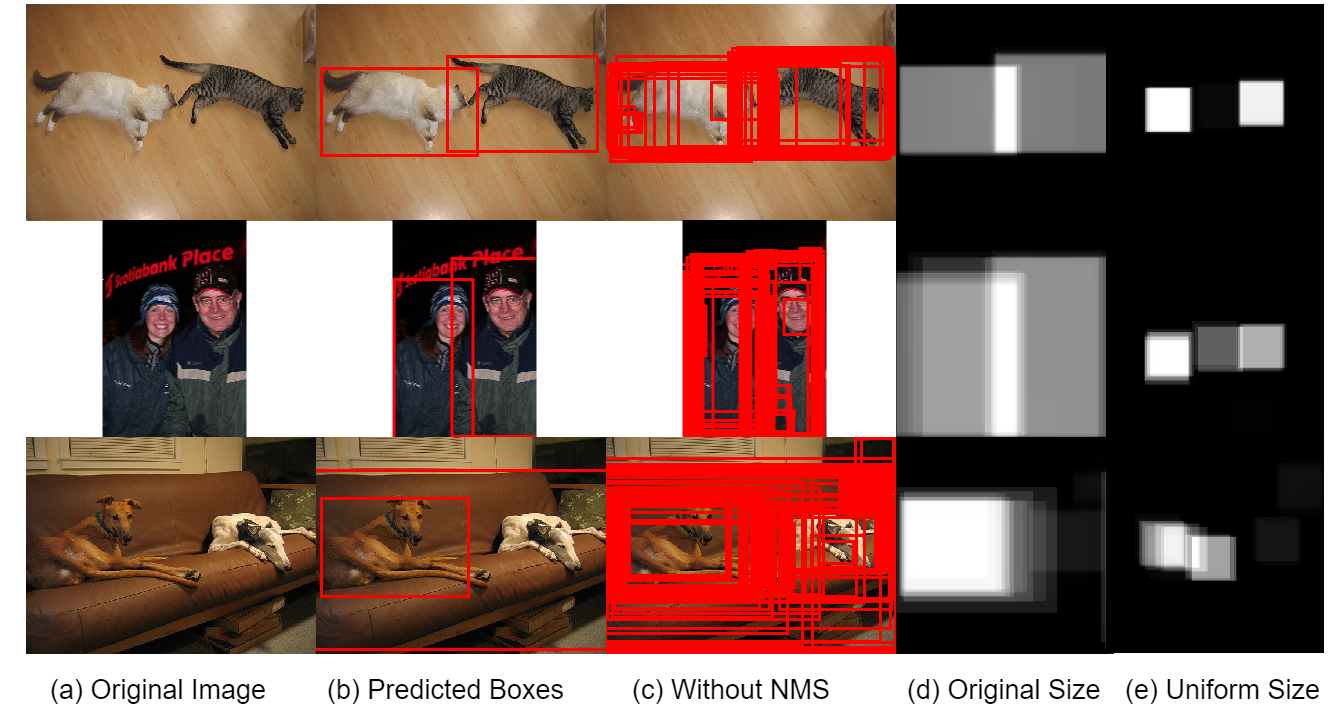}}
\end{center}
\caption{ Examples of bounding box drawn canvas images using the proposed canvas method. The first row is the training data and the second and third-row images are the test data. 
  }
\label{fig:canvas_example}
\end{figure}

 \vspace{5pt} \noindent {\bf Canvas Method} \hspace{5pt} 
In the object detection model, the model extracts numerous candidate boxes. Even for only a single object, the model predicts many predicted boxes. The NMS algorithm used in an object detection task is designed to filter out messy boxes that are predicted for a single object and predict them as a single proposed box. Using NMS, the box with the highest score is first chosen and boxes that overlap above the threshold are filtered out. To see the clear distribution of predicted boxes before the NMS, the threshold of the NMS is set to be a high value during the prediction. Because the detection model shows a different positional variance in predicting the training and test samples, this location information is important in a CNN-based attack model. In addition, similar to a classification model, the model also shows a high prediction score in the trained samples, which is crucial to a membership inference. 

To facilitate this information, we propose the use of the canvas method, which draws a predicted bounding box distribution on an empty canvas for a CNN classification network. The canvas is initially set to an image of 300$\times$300 pixels in size, where every pixel has a value of zero and the boxes drawn on the canvas have the same center as the predicted boxes and the same intensity as the prediction scores. Regarding the size of the boxes drawn on the canvas, we applied two design approaches. The first one is drawing a box equal in size as the predicted box, and the other is to draw all boxes with an identical size on the canvas regardless of the original size of the predicted box. We call the first approach the original box size, and the second the uniform box size. We use the uniform box size to make objects of all sizes detected achieve the same effect on the canvas. We set the size of a uniform box at 10\% of the canvas size. Figure \ref{fig:canvas_example} shows the examples of the canvas methods.
 
 \vspace{5pt} \noindent {\bf Augmentation} \hspace{5pt} 
 Because a bounding box distribution in a canvas image should be robust to rotations and flipping, we adopt rotation and flipping when training the attack model. We do not apply other augmentation methods such as random cropping or perspective transformation because these augmentations generate transformed bounding box distribution which might distract the target model's view on the training or test samples.

 \vspace{5pt} \noindent {\bf Score Rescaling} \hspace{5pt} 
 The prediction score of the detection model's predicted bounding box refers to how confident the model is with the objectness of bounding boxes. Because the score values are calculated after the softmax layer, the values are between zero and one. With the canvas method, bounding boxes are drawn on the canvas at the same intensity as the prediction score, and the confidence of the model might not be fully represented. For example, if the model predicts two bounding boxes with scores of 0.9 and 0.9999 respectively, it indicates that the model is much more certain that the latter is an object. However, these values do not themselves represent a significant difference on the canvas. To emphasize the model's prediction scores of the model, we utilize a score rescaling function. 

\begin{equation}\label{eq:squash_variant}
    s_{rescale} = - \log(1-s)  
\end{equation} 
In a Taylor expansion, this function is represented as $-log(1-s) =  s + \frac{s^2}{2} + \frac{s^3}{3} + ... $. Therefore in the case of an extremely small $s$, a rescale function is an  approximate identity function, which means the rescaling has little effect on small scores. Using this function, the minuscule difference between the two scores (i.e. 0.0999 from 0.9 and 0.9999) is changed to 6.91 (from 2.30 and 9.21), which can be seen as significant.

\subsection{Transfer Attack}

  We mitigate the assumption that the distribution of the target training data is similar to that of the shadow training data. In a realistic situation, it could be difficult or even impossible to secure a sufficient number of shadow data having the same distribution as the target data. Under this scenario, in \cite{ml-leaks}, a transfer attack was proposed, which composes a shadow model with relatively common and similar object detection dataset. Although a shadow model has difficulty mimicking the target model's behavior owing to different statistics and appearances between two data distributions, the attack model is still expected to be able to capture the membership status of the given data.
  
  On the other hand, the target model structure may be different. We also conducted another style of transfer attack, the shadow model structure of which differs from that of the target model. 

\section{Defense}
To mitigate a membership inference against machine learning models, we propose several defense techniques.
\subsection{Dropout}
  Because overfitting is a dominant reason why the target models leak their training data information, generalization techniques that prevent overfitting can help defend models against membership inferences. We adopt Dropout \cite{Dropout}, to obtain a well-generalized model.
\subsection{Differentially Private Algorithm}
Differential privacy \cite{dwork2011firm} offers a strong standard of privacy guarantees for computations involving aggregate datasets. It requires that any change to a single data point should reveal statistically indistinguishable differences from the model's output. A formal definition of differential privacy is described below: 
\begin{definition}[$(\epsilon , \delta) $ - Differential Privacy]\hfill

 Given two neighboring datasets $D$ and $D'$, differing by only one record, a randomized mechanism $\mathcal{A}$ provides $(\epsilon , \delta) $ - Differential Privacy  if for $\forall S \subseteq Range(\mathcal{A})$ , 
\begin{equation}
    \Pr[\mathcal{A}(D) \in S] \le e^\epsilon \Pr[\mathcal{A}(D') \in S] + \delta
\end{equation}

\end{definition}

  We call this $(\epsilon , \delta) $-DP for short. If $\delta$ = $0$ , $\mathcal{A}$ provides a stricter $\epsilon$-DP. $\epsilon$ is called a privacy loss. 
To create a differentially private deep learning model, a differentially private stochastic gradient descent (DP-SGD) \cite{abadi2016deep,mcmahan2017learning,song2013stochastic} is adopted to optimize the model. Compared to a conventional SGD optimizer, DP-SGD optimizer has two main changes to achieve the required privacy guarantee: adding Gaussian noise to gradient and gradient clipping for each minibatch sample. The specific algorithm is presented in Algorithm \ref{alg:dpsgd}. Abadi el al. \shortcite{abadi2016deep} showed a way to track a tight differential privacy bound of DP-SGD using moments accountant (MA). According to Yu et al. \shortcite{dp-publish}, however, MA assumes random sampling with replacement which is impractical and is  outperformed by random reshuffling \cite{gurbuzbalaban2015random}. Assuming sampling batches by random reshuffling, Yu et al. \shortcite{dp-publish} showed  that realistic privacy loss bound for DP-SGD is $( \rho + \sqrt{\rho \log(1/\delta)} , \delta)$-DP for $\rho$ = $\frac{k}{2 \sigma^{2}}$ where $\sigma$ is noise scale and $k$ is the number of epochs.


\begin{algorithm}[t]
\small
 \caption{Differentially Private SGD}
  \label{alg:dpsgd}
  \KwIn{Training examples $\{x_1,\ldots,x_N\}$, loss function  $\mathcal{L}(\theta) = \frac{1}{N}\Sigma_{i} \mathcal{L}(\theta,x_{i}) $,  learning rate $\eta_t$, group size $L$, noise scale $\sigma_{t}$ gradient norm bound $C$ }
  \textbf{Initialize}  $\theta_{0}$  randomly \;
  \For {$t=1~:~T$}
  {
    
    \textbf{data batching}:\\
    Take a random batch of data samples $\mathbb{B}_t$ from the training dataset\;
    $B=|\mathbb{B}_t|$\;
    \textbf{Compute gradient}:\\
    For each $i\in \mathbb{B}_t$, $\mathbf{g}_t(x_i) \leftarrow \bigtriangledown_{\theta_t} \mathcal{L}(\theta_t,x_i)$\;
    \textbf{Clip gradient}:\\
    $\hat{\mathbf{g}}_t(x_i) \leftarrow \mathbf{g}_t(x_i)/max\left(1,\frac{||\mathbf{g}_t(x_i)||_2}{C}\right)$\;
    \textbf{Add noise}:\\
    $\widetilde{\mathbf{g}}_t \leftarrow \frac{1}{B} \big(\sum_{i}\hat{\mathbf{g}}_t(x_i)+\mathcal{N}(0,\sigma_t^2C^2\mathbb{I})\big)$\;
    \textbf{Descent}:\\
    $\theta_{t+1} \leftarrow \theta_{t}- \eta_t \widetilde{\mathbf{g}}_t$ \;
    
  }
  \textbf{Output} :  $\theta_T$ \;
\end{algorithm}

\section{Experiments}
In this section, we describe the application of our method to several object detection tasks. To reduce confusion, we call the training dataset and test dataset of target and shadow models "in" and "out" data respectively. We used the Chainer framework for the object detection modules and Pytorch for the membership attack modules. 


\subsection{Target and Shadow Model Setup}
\vspace{5pt} \noindent {\bf Models} \hspace{5pt} 
To build target models, we train several object detection models including SSD and Faster R-CNN. For one-stage detection, the base SSD300-VGG16 and SSD512-VGG16 models use the VGG16 network as a backbone and have 300$\times$300 images and 512$\times$512 images as the inputs, respectively. The SSD300-Res50 model uses ResNet50 network as a backbone. For two-stage detection, the Faster R-CNN model uses the VGG16 network as a backbone and receives images with a scale of between 600 and 800.

\vspace{5pt} \noindent {\bf Datasets} \hspace{5pt} 
During the experiments, we used the datasets described earlier, i.e., VOC dataset, INRIA Pedestrian Dataset, and SynthText. According to Ahmed et al. \shortcite{ml-leaks}, one shadow dataset is sufficient. For each dataset, $D$=$(D^{train},D^{test})$, we split them by half into $(D_{target}^{in},D_{target}^{out})$ and $(D_{shadow}^{in},D_{shadow}^{out})$ to separate the target and shadow datasets. For SynthText dataset, we use the first 5,000 images with Latin characters for the target dataset and next 5,000 images for the shadow dataset.

\vspace{5pt} \noindent {\bf Training} \hspace{5pt} 
To train the SSD model, we used an SGD optimizer with an initial learning rate $10^{-3}$, $0.9$ momentum, 0.0005 weight decay and batch size 8. We trained the model 500k iterations and dropped the learning rate by $0.1$ in the 200kth, 400kth iterations. During training, we used data augmentation including horizontal flipping, color distortion, random expansion and cropping. To compare the effect of the augmentation, we also trained models that only applied flipping. To train the Faster R-CNN model, we used the same optimizer and learning rate as in the SSD and batch size 1. 

\vspace{5pt} \noindent {\bf Prediction} \hspace{5pt} 
In the case of a one-stage model, to see the overall distribution of the predicted bounding boxes, NMS threshold was set to 1.0. In the case of a two-stage model with two NMS layers, the RPN-NMS and the head-NMS thresholds were set to 0.7 and 1.0 respectively, because the high threshold value of RPN-NMS can cause a huge number of box proposals. The score threshold was set to 0.01. 
\subsection{Attack Model Setup}
To perform a black-box membership inference attack, we built several attack models as presented above. For XG-Boost model, we used Python XG-Boost package\footnote{https://github.com/dmlc/xgboost}. XG-Boost classifier takes vectorized bounding boxes and scores as inputs and has 5 maximum depth of a tree and 450 estimators as model parameters. For CNN-based classifiers with canvas method, we built two CNN models, a simple shallow CNN model and AlexNet \cite{AlexNet}. For Shallow CNN model, we used two convolutional networks having 64 and 128 channels and two fully connected networks having 128 and 2 units. CNN based attack model takes drawn canvas images with predicted boxes as input.  For balanced training, the attack model uses the almost same number of predicted results of "in" data and "out" data of the shadow model.
  We applied vertical and horizontal flipping for augmentation and score rescaling presented above. We compared various canvas methods to find the most optimal attack model.

\begin{table}[]
\begin{tabular}{p{1.1cm}|p{0.4cm}p{0.4cm}p{0.4cm}|rrrr}
\hline
Attack & \multicolumn{3}{c|}{Attack Method} & \multicolumn{2}{c|}{SSD} & \multicolumn{2}{c}{FR} \\ \cline{2-8} 
Model&Aug&BT&SR&Acc& \multicolumn{1}{l|}{AR}&  Acc&AR\\ \hline
XGB &  &  &  & 66.09 & 67.64 & 60.47 & 60.48 \\
shallow&  & (O) &  & 62.62 & 62.66 & 58.72  & 58.67  \\
AlexNet &  & (O) &  & 64.28 & 64.26 & 62.74 & 62.62 \\
AlexNet & $\checkmark$ & (O) &  &67.55  & 67.55& 64.30 & 64.22 \\
AlexNet & $\checkmark$ & (O) & $\checkmark$ & 68.30 & 68.24 & 66.59 &  66.49 \\
AlexNet & $\checkmark$ & (U) &  & 69.34 & 69.31 & 66.69 & 66.59 \\
AlexNet & $\checkmark$ & (U) & $\checkmark$ & \textbf{71.07} & \textbf{71.02} & \textbf{67.42} & \textbf{67.34}  \\ \hline
\end{tabular}
\caption{
Comparison of various attack methods. FR and XGB denote Faster R-CNN and XG-Boost. Aug, BT, SR and AR denote augmentation, box style of canvas method, score rescaling and average recall, respectively. (O) and (U) denote original and uniform box size, respectively. 
}
\label{table:attack_comparison}

\end{table}

\begin{table*}[t] 
\begin{center}
\begin{tabular}{l|c|c|r|r|r|r|r}
\multicolumn{3}{c|}{}  &\multicolumn{2}{c|}{ Target Model }  &\multicolumn{3}{c}{ Attack Model }  \\
\hline
Model  & Dataset & iters&  test mAP & train mAP & Attack Acc &Average Recall & Val Acc \\
\hline
SSD300-VGG16(Little Aug.)& VOC       & 400k  & 59.27  & 92.27  & 89.92 &  89.90 & 91.16      \\
SSD300-VGG16             & VOC       & 250k  & 73.88  & 89.30  & 67.88 &  67.92 & 72.20      \\
SSD300-VGG16             & VOC       & 500k  & 74.25  & 90.27  & 71.07 &  71.02 & 72.20      \\
SSD512-VGG16             & VOC       & 500k  & 76.53  & 91.09  & 71.03 &  70.10 & 73.04      \\
SSD300-Res50             & VOC       & 700k  & 66.04  & 85.43  & 73.86 &  73.82 & 75.97       \\
Faster R-CNN             & VOC       & 200k  & 72.71  & 88.00  & 62.50 &  62.44 & 64.44      \\
Faster R-CNN             & VOC       & 400k  & 71.80  & 90.20  & 67.42 &  67.34 & 64.44       \\
SSD300-VGG16             & INRIA     & 100k  & 88.20  & 90.90  & 71.40 &  62.95 & 73.21       \\
SSD300-VGG16             & SynthText & 400k  & 88.45  & 90.84  & 66.90 & 66.90  & 68.49      \\

\hline
\end{tabular}
\end{center}
\caption{
Attack performance on various models and datasets. Little Aug. refers training with only horizontal flipping. Attack Acc and Average recall refer attack accuracy  and average recall on target models. Val Acc refers attack accuracy on shadow models. 
}
\label{table:attack_result}
\end{table*}

\begin{figure}
\begin{center}
\subfloat{}\includegraphics[width=0.3\linewidth,height=0.25\linewidth]{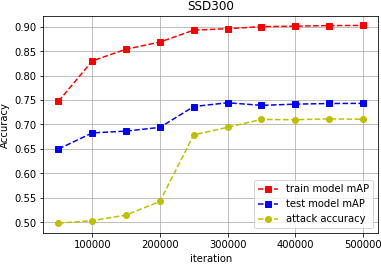}
\subfloat{}\includegraphics[width=0.3\linewidth,height=0.25\linewidth]{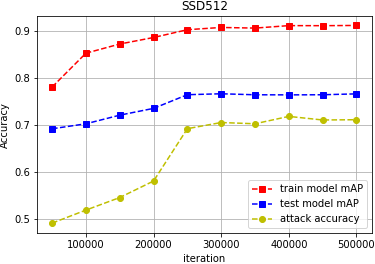}
\subfloat{}\includegraphics[width=0.3\linewidth,height=0.25\linewidth]{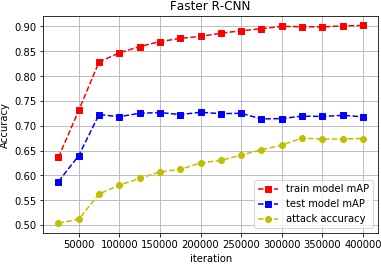}

\end{center}

\caption{Membership inference attack results on various target models.}
\label{fig:attack_graph}
\end{figure}

\subsection{Experiment Results} \label{Results_attack}

    Table \ref{table:attack_comparison} depicts the results of the comparisons of various attack methods. In general, AlexNet with augmentation, score rescaling and the uniform canvas method is successful on both the SSD and Faster R-CNN models. Therefore, we adopted the best performing method as the attack method in the next experiments.
    
    To demonstrate the relationship between the membership inference and overfitting, we conducted experiments using different numbers of iterations in the model. Figure \ref{fig:attack_graph} shows that the overall attack performance increases with an increases in the number of iterations.
    
    Table \ref{table:attack_result} shows the results of the membership inference attacks of various object detection models and datasets. The attack model is  the best performing model in table \ref{table:attack_comparison}. The mAP scores of the detection models are slightly smaller than their original performance because they train only half of the dataset. The evaluation metrics for the attack model are the accuracy and average recall of "in" and "out" labels. The attack model achieves a similar attack performance against the target and shadow models because the  distributions of dataset and model structure are similar. Overall, the attack models achieve a high accuracy for most detection of the models and datasets. In general, large generalized errors are related to the high performance of the attack models. In the case of target models trained using the INRIA and SynthText datasets, test mAP is relatively high because the tasks are easy, although the attack models still obtain a high attack accuracy.

\begin{table}[t]
\centering
\begin{tabular}{p{1cm}p{1cm}p{1cm}p{1cm}}
\hline
Model  &SSD300&SSD512&FR\\
\hline
SSD300       &  74.25 & 68.84 & 61.73   \\
SSD512       &  66.87 & 71.03 & 62.94   \\
FR           &  60.19 & 57.28 & 67.42   \\
\hline
\end{tabular}
\label{table:transfer_models}
\end{table}

\begin{table}[t]
\centering
\begin{tabular}{p{1.1cm}p{1cm}p{1cm}p{1.3cm}}
\hline
Dataset  &VOC&INRIA&SynthText\\
\hline
VOC         &  74.25 & 68.36 & 48.72       \\
INRIA       &  74.28 & 71.40 & 50.85  \\
SynthText   &  53.92 & 51.91 & 66.90  \\
\hline
\end{tabular}
\caption{Results of transfer attack over various object detection models and datasets. The x-axis represents the structure and dataset of the target models attacked and the y-axis represents that of shadow models for transfer attacks respectively. FR denotes Faster R-CNN.}
\label{table:transfer_dataset}
\end{table}

\begin{table}[t]
\centering
\begin{tabular}{p{1.6cm}p{1.3cm}p{1.5cm}p{1cm}p{1.7cm}}
\hline
Defense       &test mAP&train mAP&A-acc.&p-loss\\
\hline
Base          & 74.25  & 90.27 &  71.07 & $\infty$ \\
Dropout       &  74.20 & 89.84 & 70.94  &  $\infty$ \\
DP($\sigma$=$10^{-4}$)   & 74.32  & 88.15 & 68.68  & 2.42$\times$ $10^{10}$\\
DP($\sigma$=$10^{-3}$)    & 67.30  & 78.45 & 50.45 & 3.87$\times$ $10^{8}$  \\
\hline
\end{tabular}
\caption{Comparison of various defense methods. A-acc and p-loss denote the attack model accuracy and privacy loss respectively. }
\label{table:defense}
\end{table}

\subsection{Transfer Attacks} \label{transe}
\noindent {\bf Setup}  \hspace{5pt} 
During the transfer attack, we used the same setup as mentioned in Section \ref{Results_attack}. We conducted a  transfer attack over the SSD300, SSD512 and Faster R-CNN models and VOC dataset. We also conducted transfer attacks over VOC, INRIA, and SynthText datasets and SSD300 model.

\vspace{5pt} \noindent {\bf Results} \hspace{5pt} 
 Tables \ref{table:transfer_dataset} list the results of the model and dataset transfer attacks. The attack model trained using the same model structure or distribution dataset showed the highest accuracy. Transfer attacks on different detection models seemed to work well. The usage of the VOC dataset to attack the INRIA dataset and vice versa achieved a good performance. This might be because these two datasets have the same common label("person") and had a few objects per image. However, a transfer attack between SynthText and the other datasets did not perform well. This could be because SynthText had little in common with VOC and INRIA and had many objects per image. Transfer attacks tend to be successful when the datasets or models are similar to each other.

\subsection{Defense}

\noindent {\bf Setup}  \hspace{5pt} 
We tested the proposed defense methods against membership attacks. For the dropout, we added two dropout layers with a ratio of 0.5 before the two layers of the model. For the differentially private algorithm, we set noise scale $\sigma$=$10^{-3}$ , $10^{-4}$, gradient bound $C=50$, and minibatch size 2. We set up a relatively small noise because the object detection model has a large number of parameters \cite{general-dp}.We trained the SSD300 model 800k iterations for $\sigma$=$10^{-3}$, and 500k iterations for the others. We obtained the privacy loss with fixed $\delta$=$10^{-5}$.

\vspace{5pt} \noindent {\bf Results} \hspace{5pt} 
Table \ref{table:defense} shows the results of the defense methods. Dropout shows a slight drop in the attack accuracy, but it does not show a large difference. The DP($\sigma$=$10^{-4}$) shows little difference from the original model with mAP, but it lowers attack accuracy meaningfully. The larger noise scale DP($\sigma$=$10^{-3}$) shows some loss in accuracy, but its good defense against the attack model compensates for this.

\section{Conclusion}
In this study, we introduced new membership inference attacks against object detection models. Our proposed CNN-based attack model using the canvas method performed better than a traditional machine learning regression method. We showed that sufficiently overfitted object detection models are vulnerable to privacy leakage. A generalization error is not a guarantee of safety against an inference attack. Transfer attacks are also efficient when the models or datasets are similar. To mitigate the privacy risks, we proposed defense mechanisms that are able to reduce such risks. We showed that membership inference risks in object detection models need to be considered.

\bibliographystyle{named}
\bibliography{ijcai20}

\end{document}